\documentclass{Interspeech}

\interspeechcameraready

\title{Integrating Feedback Loss from Bi-modal Sarcasm Detector for Sarcastic Speech Synthesis}

\author[affiliation={1}]{Zhu}{Li}
\author[affiliation={1}]{Yuqing}{Zhang}
\author[affiliation={1}]{Xiyuan}{Gao}
\author[affiliation={2}]{Devraj}{Raghuvanshi}  
\author[affiliation={3}]{Nagendra}{Kumar}
\author[affiliation={1}]{Shekhar}{Nayak}
\author[affiliation={1}]{Matt}{Coler}

\affiliation{}{University of Groningen}{The Netherlands, }
\affiliation{}{Brown University}{Providence, R.I., USA \\} 
\affiliation{}{Indian Institute of Technology Indore}{Indore, India}
\email{\{zhu.li, yuqing.zhang, xiyuan.gao, s.nayak, m.coler\}@rug.nl, devraj\_raghuvanshi@brown.edu, nagendra@iiti.ac.in}


\usepackage{comment}

\begin{document}

\maketitle

\begin{abstract}
Sarcastic speech synthesis, which involves generating speech that effectively conveys sarcasm, is essential for enhancing natural interactions in applications such as entertainment and human-computer interaction. However, synthesizing sarcastic speech remains a challenge due to the nuanced prosody that characterizes sarcasm, as well as the limited availability of annotated sarcastic speech data. To address these challenges, this study introduces a novel approach that integrates feedback loss from a bi-modal sarcasm detection model into the TTS training process, enhancing the model's ability to capture and convey sarcasm. In addition, by leveraging transfer learning, a speech synthesis model pre-trained on read speech undergoes a two-stage fine-tuning process.  First, it is fine-tuned on a diverse dataset encompassing various speech styles, including sarcastic speech. In the second stage, the model is further refined using a dataset focused specifically on sarcastic speech, enhancing its ability to generate sarcasm-aware speech. Objective and subjective evaluations demonstrate that our proposed methods improve the quality, naturalness, and sarcasm-awareness of synthesized speech. 
\end{abstract}

\section{Introduction}

Sarcasm is a nuanced form of communication where speakers convey meaning contrary to their literal words, often employing exaggerated intonation, unexpected pauses, and shifts in emphasis. These subtle cues make sarcastic speech challenging to interpret and synthesize \cite{li2024functional, cheang2009acoustic}. As speech technology increasingly integrates into everyday applications, such as virtual assistants, interactive media, and conversational AI, the ability to accurately generate colloquial and expressive speech, including sarcasm, is crucial for enhancing human-computer interaction.

Text-to-speech (TTS) systems have advanced significantly with the advent of deep learning, enabling high-quality, natural-sounding speech synthesis from text \cite{tan2021survey, wang2017tacotron, ren2019fastspeech}. However, most existing systems focus on generating neutral or reading-style speech, with limited success in improving speech synthesis in spontaneous or dynamic contexts or capturing expressive variations such as sarcasm \cite{ping2017deep, shen2018natural, ren2020fastspeech}. 
Recent research has sought to address these challenges by introducing methods to capture and control emotional expressiveness and improve the prosody and variability of synthetic speech. For instance, Li et al. \cite{li2021controllable} conduct controllable emotional transfer by incorporating an emotion style classifier with a feedback loop, where the classifier encourages the TTS model to generate speech with specific emotions. Similarly, O’Mahony et al. \cite{o2022combining} combine real spontaneous conversations with read speech to augment the training data to improve the prosody of synthetic speech. Li et al. \cite{li2023towards} achieve high-quality expressive speech synthesis by introducing a semi-supervised pre-training approach that leverages a large-scale, low-quality spontaneous speech dataset. This method enriches both the quantity of spontaneous-style speech and the diversity of associated behavioral labels, improving the naturalness and expressiveness of the synthesized speech.

Despite significant advancements in enhancing the expressiveness and prosody of TTS systems, synthesizing more complex forms of speech, such as sarcasm, remains a challenging task. Sarcasm often emerges in spontaneous, conversational contexts, where the subtle and dynamic nature of human speech is most pronounced, making its synthesis particularly difficult.
Sarcasm, with its distinctive expressive qualities—such as exaggerated intonation, unexpected pitch variations, and altered timing—requires a more nuanced approach to accurately model the dynamic nature of speech. These unique features of sarcastic speech present a challenge for current TTS models, which are primarily trained on neutral or standard emotional speech.

Another major obstacle to progress in sarcastic speech synthesis is the limited availability of dedicated datasets \cite{li2025leveraging}. Sarcasm in speech is less frequently recorded and studied compared to neutral or emotional speech, limiting the training data available for TTS models.
The scarcity of training data restricts the ability of models to learn and accurately reproduce the specific features of sarcasm. 
While techniques such as data augmentation and transfer learning have been explored to mitigate the lack of data \cite{li23_ssw, huybrechts2021low, tits2020exploring}, they have yet to fully overcome the inherent difficulties in capturing the essence of sarcasm in synthesized speech.
For instance, Huybrechts et al. \cite{huybrechts2021low} introduced a novel 3-step methodology to create expressive style voices with minimal data. Their approach leverages voice conversion to augment data from other speakers, followed by training a TTS model on both the augmented and available recordings, and fine-tuning to further enhance quality. This method demonstrates the potential of data augmentation techniques to build expressive TTS systems with minimal data. However, despite such advancements, synthesizing complex emotional speech like sarcasm remains a significant challenge.

In contrast to the relatively limited work on sarcastic speech synthesis, the field of sarcasm detection has received considerable attention, with approaches that utilize both unimodal (e.g., audio) and multimodal (e.g., audio, visual, and textual) data \cite{tepperman2006yeah, rakov2013sure, raymultimodal, gao2024amused, raghuvanshi2025intra}.
In audio-based approaches, machine learning methods have predominated, with the selection of acoustic features such as prosodic, spectral, and contextual cues playing a crucial role \cite{tepperman2006yeah, rakov2013sure}. 
Multimodal approaches leads to the introduction of several multimodal sarcasm datasets that encompass audio, visual, and textual modalities \cite{raymultimodal, gao2024amused, raghuvanshi2025intra, gao2022deep, castro2019towards}, on which detection benchmarks have been established. 
Subsequent research has concentrated on refining methods for integrating these different modalities and improving fusion techniques to enhance detection accuracy \cite{Schifanella, Wu}. 
Sarcasm detection models have demonstrated significant potential in identifying the acoustic and contextual markers of sarcasm, and the insights from these models could be valuable for guiding sarcastic speech synthesis. 

However, to date, there has been limited exploration of how sarcasm detection could inform the generation of sarcastic speech in TTS systems.
Building on advances in multi-modal sarcasm detection and drawing on the success of data augmentation techniques in low-resource TTS, we propose three methods to enhance the sarcasm-awareness of the sarcastic speech synthesis model and improve the quality and naturalness of the synthesized speech.
 
The main contributions of this work are summarized as follows:

\begin{enumerate}
    \item \textbf{Novel Integration of Sarcasm Detection Feedback:} We introduce a novel approach that integrates a bi-modal sarcasm detector into the sarcastic speech synthesis pipeline. Specifically, we incorporate feedback loss derived from sarcasm detection into the TTS model training process. This integration enhances the model's ability to capture the nuances and expressiveness of sarcastic speech, making it sarcasm-aware. Both objective and subjective evaluation results demonstrate that this incorporation enhances the model's ability to convey sarcasm, enabling it to better capture the subtle and expressive elements of sarcastic speech. 
    \item \textbf{Comprehensive Comparison of Input Modalities:} We perform a comprehensive comparison of various input types for sarcasm detection, revealing that combining text and speech inputs leads to a notable improvement in the F1-score. This finding highlights the importance of incorporating textual information to detect sarcastic behaviors that are difficult to identify through speech alone, underscoring the complexity of sarcasm.
    \item \textbf{Two-Stage Fine-tuning Method:} We propose a two-stage fine-tuning method for TTS models, which improves the model's performance in generating sarcastic speech. This approach effectively addresses the challenge of limited sarcastic speech data, guiding the model toward producing more accurate and appropriate sarcastic prosody.
\end{enumerate}

\section{Methodology}

This section details the key elements of the proposed model, including a description of the bi-modal sarcasm detection model and the data augmentation method employed in a two-stage fine-tuning approach.

\begin{figure}[h]
    \centering
    \includegraphics[width=1\linewidth]{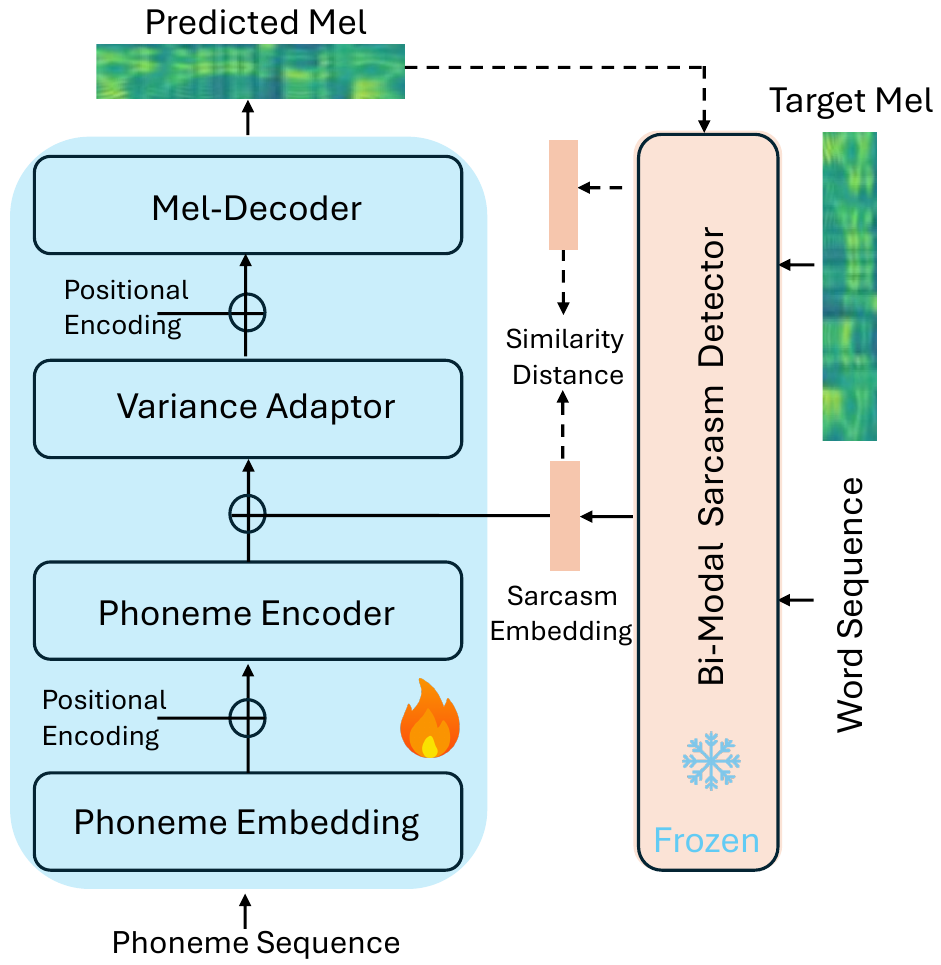}
    \caption{The model architecture for sarcastic speech synthesis.}
    \label{fig:proposed}
\end{figure}

\begin{figure}
    \centering
    \includegraphics[width=1\linewidth]{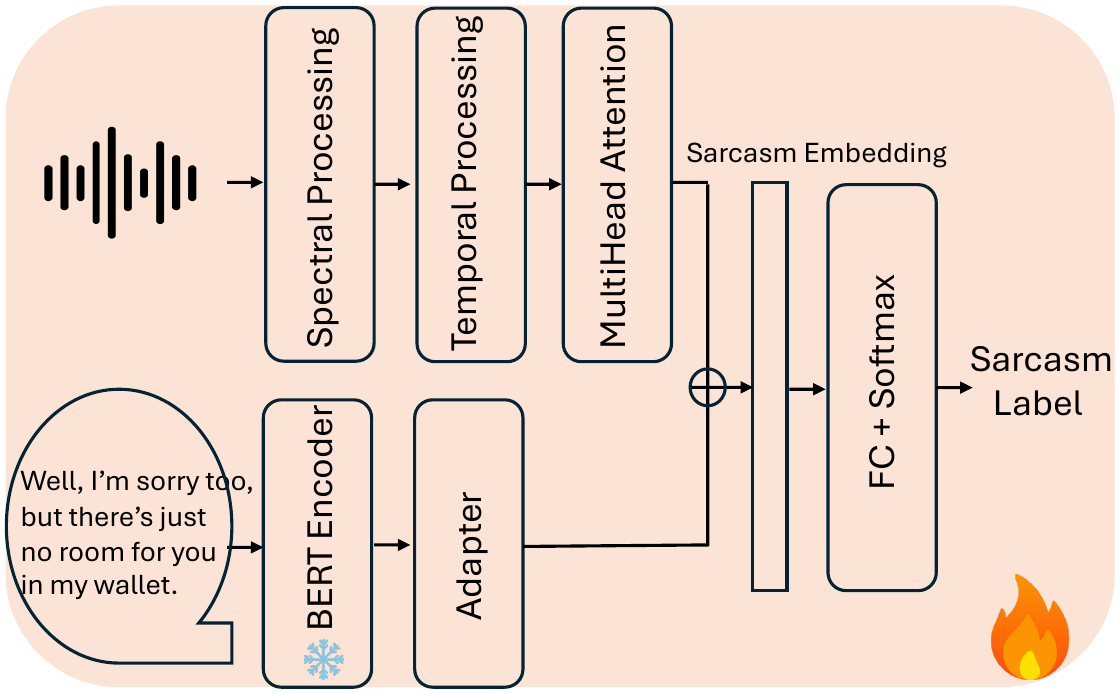}
    \caption{The structure of the bi-modal sarcasm detector.}
    \label{fig:detector}
\end{figure}

\subsection{Overall Model Architecture}
The architecture of our proposed sarcastic speech synthesis model is illustrated in Fig. \ref{fig:proposed}. The backbone of the TTS model integrates FastSpeech 2 \cite{ren2020fastspeech} to generate mel spectrograms from input phoneme sequences. To model and predict sarcastic speech, we incorporate a bi-modal sarcasm detector that extracts sarcasm embeddings, enabling the generation of appropriate speech (i.e., sarcastic or non-sarcastic). This injects sarcasm-related features into the TTS model, effectively making it sarcasm-aware. By providing the model with prior knowledge of sarcasm, the sarcasm embeddings guide the speech synthesis process to generate appropriate prosody and tone.

Specifically, our proposed model training procedure consists of the following two steps: 
\begin{itemize}
    \item Training the Sarcasm Detector. We train an independent detection model for the sarcasm labels using a labeled sarcastic dataset.
    \item Two-stage Fine-tuning. Since sarcastic speech is inherently conversational, the pre-trained TTS model is fine-tuned on a conversational dataset to learn diverse speech styles. Then, in the second stage, the model is fine-tuned using a labeled sarcastic dataset to enhance sarcasm generation. 
\end{itemize}
We detailed the training procedure in the following sections \ref{sec:detector} and \ref{combined_data}.

\subsection{Integration of a Bi-modal Sarcasm Detector}
\label{sec:detector}

To effectively synthesize sarcastic speech, it is crucial that the synthesized output accurately reflects the sarcastic tone. Incorporating a sarcasm detector into the speech synthesis pipeline can make the model sarcasm-aware. 
Detecting sarcasm in speech based solely on acoustic features is challenging, especially when using low-quality, conversational sarcastic datasets \cite{raymultimodal}. In addition, sarcasm is often conveyed through specific semantic cues and the presence of specific words. To address this, we constructed a bi-modal sarcasm detector, inspired by the structure in \cite{nielsen2020role}, which leverages text as auxiliary information.

The proposed bi-modal sarcasm detector (illustrated in Fig.\ref{fig:detector}) begins by extracting mel spectrograms from the speech data. The mel spectrograms are subsequently processed through spectral processing, temporal processing, and multi-head self-attention \cite{min2021meta}. The obtained features are concatenated with word embeddings derived from the text \cite{devlin2018bert}. The combined features are fed into a fully connected layer to predict sarcasm labels. By integrating both modalities, the detector can more accurately capture the nuanced nature of sarcasm in speech.

The bi-modal sarcasm detector is integrated into the TTS pipeline as follows: During training, the input text and the target speech are passed through the sarcasm detector to generate a sarcasm embedding. This embedding is then concatenated with the phoneme encoder output before being fed into the variance adaptor. In this way, sarcasm-related features are injected into the TTS model. During inference, the sarcasm detector generates an embedding based on the input text and the reference speech, which is used similarly to guide the speech synthesis process.


For the loss function, in addition to the mean absolute error (MAE) between predicted mel spectrograms and the ground truth spectrogram, we use the cosine distance between the ground truth audio+text embedding and the one extracted from the predicted Mel-spectrogram+text embedding by the bi-modal sarcasm detector as one of the loss functions for optimizing the TTS network.

\subsection{Two-stage Fine-tuning}
\label{combined_data}
    
Given the scarcity of sarcastic datasets for speech synthesis, we address this challenge by employing a two-stage fine-tuning approach, which enables our TTS model to learn from diverse speech styles and capture the nuances of sarcasm.

In the first stage, the model is pre-trained on a high-quality read-style speech dataset. This stage serves as the foundation for the model, where it learns to generate clear, natural-sounding speech with accurate prosody. Read speech datasets are commonly used for initial training because of their high quality and consistency, making them ideal for establishing the core speech synthesis capabilities. By starting with this robust baseline, the model is able to generate neutral, well-articulated speech that will serve as a solid foundation for subsequent fine-tuning.

The second stage focuses on data augmentation and fine-tuning using a curated conversational speech dataset. Conversational speech encompasses a broader range of speech patterns and prosody, such as varying intonations, pacing, and emotional expressiveness, which are not as prominent in neutral read-style speech. Fine-tuning the model on this dataset helps introduce more variability into the model's output, enabling it to generate more dynamic and context-aware speech. This step effectively enhances the model's ability to handle diverse speech styles, making it better suited to more expressive speech forms.

Finally, in the third stage, the model is fine-tuned again using a labeled sarcastic dataset. Sarcasm requires a model to capture subtle intonational cues, pitch variations, and context-dependent exaggerations, which differ significantly from standard emotional speech. By incorporating a labeled sarcastic dataset, the model can learn to recognize and reproduce these specific features. This fine-tuning step is crucial for adapting the model to the unique demands of sarcastic speech synthesis, where the tone, timing, and delivery need to be precisely aligned with the intended sarcastic meaning.

This two-stage fine-tuning process allows the model to progressively improve its ability to generate speech that spans from neutral to highly expressive, including the complex and nuanced patterns found in sarcasm.

\section{Experiments}

\subsection{Data}

To build our sarcastic text-to-speech (TTS) system, we employ a multi-stage training approach utilizing several carefully selected datasets.

\textbf{Pre-training.}
For the initial pre-training phase, we use the LibriTTS corpus \cite{zen2019libritts}, a high-quality, read-style English speech dataset derived from audiobooks. LibriTTS provides diverse speakers and clean audio recordings, making it a strong foundation for building a robust baseline TTS model.

\textbf{Fine-tuning Stage I: Conversational Speech Adaptation.}
Given that sarcastic speech typically arises in casual, conversational settings, the first stage of fine-tuning adapts the TTS model to natural dialogue-style speech. We compile an auxiliary dataset of conversational utterances by extracting audio from episodes of Friends and The Big Bang Theory, aligning with the domains used in the MUStARD++ dataset. The raw audio is processed using Emilia-Pipe \cite{he2024emilia}, an open-source preprocessing pipeline tailored for in-the-wild speech data. Emilia-Pipe conducts standardization, music and noise separation, speaker diarization, voice activity detection (VAD)-based segmentation, automatic speech recognition (ASR), and filtering. This pipeline ensures high-quality, segment-level audio data with precise textual alignment, yielding a total of 6.17 hours of augmented conversational speech suitable for TTS training.

\textbf{Fine-tuning Stage II and Sarcasm Detection.}
For the final stage of fine-tuning and for training of the sarcasm detector, we use the sarcasm dataset, A Multimodal Corpus for Emotion Recognition in Sarcasm (MUStARD++) \cite{raymultimodal}. MUStARD++ is a multimodal sarcasm detection corpus compiled from popular sitcoms such as Friends and The Big Bang Theory. It contains 1,202 audiovisual utterances, equally divided into 601 sarcastic and 601 non-sarcastic samples. Each utterance is paired with its conversational context—preceding utterances in the dialogue—providing valuable contextual cues for detecting sarcasm.

In the preprocessing phase, we resample all audio to 22,050 Hz for consistency with our TTS model. To enhance audio quality and improve model performance, we remove silence at the beginning and end of each clip. Since sarcastic speech often includes overlapping speech and laughter, we use Demucs \cite{defossez2019demucs}, a state-of-the-art music source separation tool, to isolate the human voice from background laughter, music, and ambient noise.

For evaluation, we randomly select 100 utterances as the test set, shared across both the sarcasm detection model and the sarcastic TTS system to maintain consistency in performance comparison.

\subsection{Compared Methods}

To evaluate the effectiveness of the proposed bi-modal sarcasm detector, which forms a crucial component of the synthesis pipeline, we compare it against the baseline sarcasm detector used in MUStARD++ \cite{raymultimodal}.
To evaluate the effectiveness of our proposed sarcastic TTS system, we compare it against a baseline TTS model. Both models are built upon the FastSpeech 2 architecture \cite{ren2020fastspeech}. The models compared are as follows:

\textbf{Baseline Sarcasm Detector.}
We followed MUStARD++ \cite{raymultimodal} for extracting features and building sarcasm detection systems. Unlike the original implementation of the MUStARD++ framework\footnote{https://github.com/cfiltnlp/MUStARD\_Plus\_Plus}, we only use two modalities: text and audio.
For feature extration from the text modality, we encode the text using BERT \cite{devlin2018bert} with $d_{t} = 768$ and use the mean of the last four transformer layer representations to get a unique embedding representation for each utterance. For audio modality, we extract MFCC, Mel spectrogram and prosodic features of size $d_{m}$,  $d_{s}$, $d_{p}$ respectively. Then we take the average across segments to get the final feature vector. Here $d_{m} = 128$,  $d_{s} = 128 $, $d_{p} = 35 $ , so our audio feature vector is of size $d_{a} = 291 $.

\textbf{Proposed Sarcasm Detector.}
As described in Section~\ref{sec:detector}, the proposed bi-modal sarcasm detector first extracts mel spectrograms from speech, which are processed through spectral and temporal modules followed by multi-head self-attention \cite{min2021meta} to capture acoustic patterns. Simultaneously, BERT \cite{devlin2018bert} is used to encode the text into semantic embeddings. The resulting speech and text features are concatenated and passed through a fully connected layer to predict sarcasm labels, effectively leveraging both prosodic and contextual cues.
The feature extraction parameters remain largely consistent with the baseline, except that we use only mel spectrograms for the audio modality.

\textbf{Baseline Sarcasm Synthesis Model.}
The baseline is an open-source implementation of the standard FastSpeech 2 model \footnote{https://github.com/ming024/FastSpeech2}. The model does not incorporate any sarcasm-specific cues, labels, or fine-tuning strategies, making it a strong reference point for evaluating the impact of our proposed enhancements.

\textbf{Proposed  Sarcasm Synthesis Model.}
Our proposed system extends FastSpeech 2 through a two-stage fine-tuning pipeline combined with a bi-modal sarcasm detection feedback mechanism. The first fine-tuning stage adapts the base model to conversational speech using sitcom-derived data. The second stage incorporates sarcastic speech from the MUStARD++ dataset to guide the model toward generating sarcastic prosody and intonation. Additionally, we integrate a sarcasm detector during training as a feedback constraint, encouraging the synthesized speech to not only sound natural but also be perceived as sarcastic. This bi-modal approach leverages both acoustic and textual cues, making the generation more context-aware and emotionally expressive.

\subsection{Model Configuration}

The baseline TTS model configuration and hyperparameters follow the original FastSpeech 2 implementation \cite{ren2020fastspeech}. The model architecture includes 4 feed-forward Transformer (FFT) blocks in both the encoder and the mel-spectrogram decoder. In each FFT block, the dimension of phoneme embeddings and the hidden size of the self-attention are set to 256. The combine layer utilizes a 1D convolutional network with ReLU activation, featuring an input size of 1024 and an output size of 256. The decoder's output linear layer transforms the hidden states into 80-dimensional mel spectrograms. The phoneme duration is extracted by Montreal Forced Aligner tool \cite{mcauliffe2017montreal}. In the training bi-modal sarcasm detector phase, we train 50 epochs with a batch size set to 256. For the TTS model, we perform 800k iterations for pre-training and 100k iterations for two-stage fine-tuning. The Adam optimizer is employed with hyperparameters $\beta_1 = 0.9$, $\beta_2 = 0.98$, and $\varepsilon = 10^{-9}$. The generated mel spectrograms are subsequently converted into waveforms using the HiFi-GAN vocoder \cite{kong2020hifi}.

\section{Results and Discussion}

In this section, we present an evaluation of our sarcastic speech synthesis model. We begin by assessing the performance of our bi-modal sarcasm detector, which forms a crucial component of the synthesis pipeline. We then evaluate the quality of the synthesized speech through objective metrics and subjective human evaluations. In addition, the proposed model is compared against the baseline model using a preference test.
The evaluation results demonstrate the effectiveness of incorporating sarcasm detection feedback and our two-stage fine-tuning approach in generating more natural and appropriate sarcastic speech. Speech samples are available at \url{https://abel1802.github.io/SarcasticTTS/}. 

\subsection{Sarcasm Detection Performance}

We conducted an objective evaluation to assess the performance of our proposed bi-modal sarcasm detector using precision, recall, and F1-score as evaluation metrics. The baseline for comparison is the default sarcasm detector used in MUStARD++ \cite{raymultimodal}. Table \ref{tab:detection_result} presents a comparative analysis of detection performance across different input types: 1) Speech-only, where the detector relies solely on acoustic features; 2) Speech+Text, where the detector uses both audio and transcribed text for joint inference.
The results show that our proposed detector consistently outperforms the baseline across both input types. Notably, the bi-modal configuration (speech+text) demonstrates a substantial gain in F1-score, improving from 68.7\% (MUStARD++) to 71.2\% (ours). This indicates that integrating textual information provides important semantic cues, aiding in detecting sarcastic behaviors that may not be as evident in speech alone. 

\begin{table}[h!]
  \caption{Sarcasm detection on real data}
  \label{tab:detection_result}
   \centering
       \resizebox{\columnwidth}{!}{
  \begin{tabular}{ccccc}
    \toprule
    \textbf{Method} & \textbf{Input type} & \textbf{Precision (\%)} & \textbf{Recall (\%)} & \textbf{F1-score (\%)}
    \\ 
    \midrule
    MUStARD++ & speech & 63.9 & 63.5 & 63.6
    \\
    Proposed detector & speech & 66.6 & 66.3 & 66.2
    \\
    MUStARD++ & speech+text & 68.8 & 68.6 & 68.7
    \\ 
    Proposed detector & speech+text & 71.3 & 71.2 & 71.2
    \\ 
    \bottomrule

  \end{tabular}}
\end{table}

These results confirm the advantage of our bi-modal architecture, which effectively captures sarcasm's nuanced expression by leveraging both prosodic (e.g., pitch, rhythm, intensity) and semantic (e.g., word choice, context) cues. The performance gains further validate the utility of integrating such a detector into the TTS training process.

Building on this strong performance, we incorporate the sarcasm detector into our TTS framework (as described in Section~\ref{sec:detector}) to infuse sarcasm-awareness into the speech synthesis process. In the next section, we evaluate the synthesized speech to determine the extent to which this integration improves the naturalness and sarcastic quality of the generated audio.

\subsection{Objective Evaluation of Sarcasm TTS}

To objectively assess the sarcasm expressivity of our proposed TTS model, we evaluated the ability of the sarcasm detection model to recognize sarcasm in speech synthesized by both the baseline and our proposed model. Table \ref{tab:detection_tts_result} presents the performance of sarcasm detection under two input conditions for the detector: (1) speech-only and (2) speech combined with the original text (speech+text).

In the speech-only condition, both models achieve comparable performance, with the proposed model slightly outperforming the baseline (F1-score of 63.4\% vs. 62.2\%). This modest improvement suggests that, even without access to textual context, the prosodic and acoustic features generated by the proposed model better capture the subtleties of sarcastic speech. This is likely due to the enhanced modeling of expressive speech patterns during training.

The performance gains become more prominent in the speech+text condition, where the sarcasm detector leverages both the audio and the corresponding textual content. Here, the proposed model achieves the highest performance with an F1-score of 70.1\%, surpassing the baseline’s score of 68.8\%. This improvement in both precision and recall indicates that our model’s synthesized speech aligns more closely with the sarcastic cues embedded in the accompanying text. In other words, the synthesized prosody and speech characteristics are more semantically consistent with the sarcastic intent expressed in the text.


These results clearly demonstrate the advantage of bi-modal learning. By jointly modeling speech and text during training, the proposed system learns to generate audio that not only sounds natural but also faithfully conveys complex communicative intents like sarcasm. The sarcasm detector's improved performance on our synthesized samples serves as indirect but objective evidence that sarcasm is more effectively encoded in our generated speech.

In summary, these findings support the conclusion that our sarcasm-aware TTS model is better equipped to generate expressive and contextually appropriate speech, especially when paired with textual context. The integration of the bi-modal sarcasm detection mechanism within the training loop proves important in enhancing the expressiveness of the synthesized output.

\begin{table}[h!]
  \caption{Sarcasm detection on generated data}
  \label{tab:detection_tts_result}
   \centering
    \resizebox{\columnwidth}{!}{
  \begin{tabular}{ccccc}
    \toprule
    \textbf{Method} & \textbf{Input type} & \textbf{Precision (\%)} & \textbf{Recall (\%)} & \textbf{F1-score (\%)}
    \\ 
    \midrule
    Baseline & speech & 61.5 & 66.7 & 62.2
    \\
    Proposed & speech & 61.9 & 68.6 & 63.4
    \\ 
    Baseline & speech+text & 68.9 & 68.7 & 68.8
    \\ 
    Proposed & speech+text & 71.4 & 69.8 & 70.1
    \\
    \bottomrule

  \end{tabular}}
\end{table}

\subsection{Subjective Evaluation of Sarcasm TTS}

\begin{figure}[h]
    \centering
    \includegraphics[width=1\linewidth]{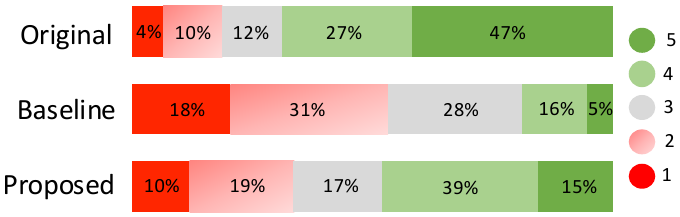}
    \caption{This figure indicates the percentage of participants who provided ratings 1 (the lowest) to 5 (the highest) for each model.}
    \label{fig:mos}
\end{figure}

\begin{figure}[h]
    \centering
    \includegraphics[width=1\linewidth]{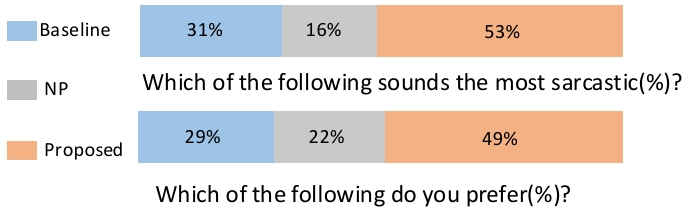}
    \caption{Result for preference test. NP means no preference.}
    \label{fig:pref}
\end{figure}

Both the Mean Opinion Score (MOS) evaluation (Figure~\ref{fig:mos}) and preference tests (Figure~\ref{fig:pref}) were conducted to assess listeners' perceptions of the sarcastic speech generated by each model. Thirteen listeners with no reported hearing impairments participated in this evaluation. The results indicate that the proposed model achieved a significantly higher MOS compared to the baseline. Notably, 15\% utterances generated by the proposed model were assigned the highest possible score, compared to only 5\% for the baseline model, demonstrating a clear preference for the sarcasm-aware synthesis.

In the preference tests, 53\% of utterances generated by the proposed model were rated as having a stronger sarcasm tendency, while 49\% of utterances generated using the proposed model were preferred. Listeners frequently rated it as more natural and sarcastic, highlighting the model's ability to capture the nuanced expressiveness of sarcasm. Conversely, the baseline model received the lowest preference, with many participants describing its sarcastic speech as less convincing and somewhat monotonous. These results highlight the superior performance of the proposed model in generating natural sarcasm and suggest its potential for further refinement in sarcastic speech synthesis.
In summary, these results demonstrate that 
our model improves upon the baseline FastSpeech 2 in generating sarcastic speech.
The integration of sarcasm detection feedback and our two-stage fine-tuning approach contribute to the generation of more natural sarcastic speech, as evidenced by objective metrics and subjective evaluations.

\section{Conclusion}
This study presents the first comprehensive approach to sarcastic speech synthesis, introducing a novel integration of sarcasm detection feedback into the TTS training pipeline. By incorporating a bi-modal sarcasm detector as an auxiliary guidance mechanism and leveraging a two-stage fine-tuning process, our method improves the expressiveness and naturalness of synthesized sarcastic speech. Both objective metrics and subjective listening tests demonstrate that our model more effectively captures the unique prosodic and semantic features characteristic of sarcasm.


A key strength of this work lies in its pioneering focus on sarcastic speech, an underexplored and complex emotional style that is challenging to model due to its subtlety and contextual dependence. By drawing inspiration from advances in multimodal sarcasm detection, this study bridges a gap between detection and synthesis tasks, offering a promising direction for expressive and emotionally intelligent speech generation. Additionally, the use of sitcom-derived conversational speech and sarcasm-labeled datasets ensures ecological validity, reflecting real-world sarcastic interactions.



Despite these advancements, our study has some limitations. Most notably, we do not isolate the individual contributions of the sarcasm detector and two-stage fine-tuning. Such an analysis would offer deeper insight into which components contribute most to performance improvements and where further optimization might be focused. 
Another limitation of this study is the absence of a control condition with neutral, non-sarcastic texts. All evaluation samples were drawn from a sarcasm dataset, which may have primed listeners toward sarcasm, regardless of the actual prosodic cues. As a result, it is difficult to disentangle whether listeners' sarcasm judgments were based on the synthesized speech itself or influenced by the sarcastic nature of the text. This raises concerns about what listeners were truly responding to—whether it was the intended prosody, overall speech quality, or merely the sarcastic content of the text.

Future research should aim to expand and diversify the dataset to include both sarcastic and neutral texts, enabling more controlled evaluations. Incorporating fine-grained modeling of sarcasm, such as different types or degrees of sarcastic expression, could lead to more nuanced synthesis. Additionally, further efforts are needed to develop methods that more effectively integrate contextual information, both linguistic and situational. Finally, refining evaluation metrics to better capture perceptual subtleties, such as listener sensitivity to sarcasm versus general speech quality, will be essential for advancing the field of sarcastic speech synthesis.


\section{Acknowledgement}
We would like to thank the Mentoring Experiences for Underrepresented Young Researchers (ME-UYR) program, sponsored by the IEEE Signal Processing Society (SPS), for its valuable support and mentorship, which significantly contributed to the development of this work.

\bibliographystyle{IEEEtran}
\bibliography{mybib}

\end{document}